\documentclass[letterpaper]{article}
\usepackage{aaai}
\usepackage{times}
\usepackage{helvet}
\usepackage{courier}
\usepackage{multirow}
\usepackage{multicol}
\usepackage{booktabs}
\usepackage{amsmath}
\usepackage{amssymb}
\usepackage{color}

\newcommand{\citet}[1]{\citeauthor{#1} \shortcite{#1}}
\newcommand{\citep}{\cite}

\usepackage{prettyref,soul}

\newcommand{\nosection}[1]{\vspace{2pt}\noindent\textbf{#1.}}
\newtheorem{theorem}{\textbf{Theorem}}
\newtheorem{lemma}{Lemma}
\newtheorem{property}{Property}
\newtheorem{assumption}{Assumption}
\newtheorem{proof}{Proof}
\begin{document}
%
\title{Characterizing Membership Privacy in Stochastic Gradient Langevin Dynamics}
\author{{\bf Bingzhe Wu}\textsuperscript{1}, {\bf ChaoChao Chen}\textsuperscript{2}, {\bf Shiwan Zhao}\textsuperscript{3}, {\bf \Large Cen Chen}\textsuperscript{2}, \\{\bf \Large Yuan Yao}\textsuperscript{4}, {\bf \Large Guangyu Sun}\textsuperscript{1}, {\bf \Large Li Wang}\textsuperscript{2}, {\bf \Large Xiaolu Zhang}\textsuperscript{2}, {\bf \Large Jun Zhou}\textsuperscript{2}\\ \textsuperscript{1}Peking University, \textsuperscript{2}Ant Financial Service Group, \textsuperscript{3}IBM Reseach, \textsuperscript{4}HongKong University of Science and Technology}
\maketitle
\begin{abstract}
Bayesian deep learning is recently regarded as an intrinsic way to characterize the weight uncertainty of deep neural networks~(DNNs). Stochastic Gradient Langevin Dynamics~(SGLD) is an effective method to enable Bayesian deep learning on large-scale datasets. Previous theoretical studies have shown various appealing properties of SGLD, ranging from the convergence properties to the generalization bounds. 

In this paper, we study the properties of SGLD from a novel perspective of membership privacy protection (i.e., preventing the membership attack). The membership attack, which aims to determine whether a specific sample is used for training a given DNN model, has emerged as a common threat against deep learning algorithms. 
To this end, we build a theoretical framework to analyze the information leakage (w.r.t. the training dataset) of a model trained using SGLD. Based on this framework, we demonstrate that SGLD can prevent the information
leakage of the training dataset to a certain extent. Moreover, our theoretical analysis can be naturally extended to other types of Stochastic
Gradient Markov Chain Monte Carlo (SG-MCMC) methods. Empirical results on different datasets and models verify our theoretical findings and suggest that the SGLD algorithm can not only reduce
the information leakage but also improve the generalization ability of the DNN models in real-world applications.

\end{abstract}
\section{Introduction}
Bayesian deep learning has received increasing attention from the research community over the past few years, due to its effectiveness to capture weight uncertainty and alleviate overfitting in deep neural networks~(DNNs)~\cite{BBB,psgld}. The key ingredient of Bayesian deep learning is sampling from the Bayesian posterior. 
To enable the posterior sampling, there are typically two different types of approaches, namely, Stochastic Variational Inference~(SVI) and Stochastic Gradient Markov Chain Monte Carlo (SG-MCMC).
One representative work of SVI is Bayes-by-Backprop~\cite{BBB} that enables the variational inference of DNNs. 
These SVI based methods in the realm of deep learning always assume that the Bayesian posterior can be decomposed into separable Gaussian distributions, which may lead to non-negligible approximation error and underestimation of the weight uncertainty~\cite{psgld}. 
Such SVI based approaches can hardly scale to large datasets and difficult to handle complex model structures, thus limit their applicability for deep learning models.

To enable Bayesian learning at scale, SG-MCMC is proposed as an alternative to SVI~\cite{sgld_original,SGHMC}. 
Stochastic Gradient Langevin Dynamics (SGLD) is one of the most popular SG-MCMC methods.
The key idea of SGLD is applying stochastic optimization~\cite{stochastic_optimization} to Langevin Dynamics~\cite{langevin_dynamics}, which is implemented by injecting proper Gaussian noise to the gradient estimation of a mini-batch sampled from the training dataset~\cite{sgld_original}. SGLD can also be regarded as an effective way to approximately sample from the Bayesian posterior distribution~\cite{sde_view_sgld}. 
It has been successfully applied to a number of real-world applications, such as language modeling~\cite{sgld_language_model} and shape 
classification~\cite{sgld_shape_classification}. Among these practical applications, various theoretical studies have shown different appealing properties of SGLD \cite{sde_view_sgld,colt_sgld_gen}. Most of these theoretical works focus on analyzing the convergence properties of SGLD. 
Besides the convergence properties, some recent works 
also attempted to study the generalization bounds of SGLD~\cite{colt_sgld_gen,ISIT_sgld_gen}. 
 
In this paper, we specifically study the properties of SGLD from the perspective of membership privacy.
Some recent studies showed that a well-trained DNN model may unintentionally expose the private information of the training dataset~\cite{ccs_model_inversion,s_and_p_attack}. 
The membership attack~\cite{s_and_p_attack} has emerged as a common threat to detect the membership information of the training dataset armed with the learned model. 
Given the trained model and a specific sample, the membership attack aims to infer whether this given sample is used for training the given model. Our work is motivated by some recent works that study the connection between the overfitting and the information leakage of DNNs, which shows that the less overfitting a model is, the less information is disclosed by the model~\cite{wu2019p3sgd,overfitting_privacy,GAN_privacy}.

To this end, we first build the theoretical framework to analyze the information leakage of the learned model. Specifically, we employ the optimal membership attack proposed in a previous work~\cite{optimal_attack_ICML} as a workhorse to compute the membership privacy leakage loss. This loss can be used for quantitatively characterizing the model's vulnerability to the membership attack. Based on the theoretical framework, we then prove that, for a model trained using SGLD, its membership privacy leakage loss can be bounded in expectation by a uniform constant for arbitrary samples. This finding further indicates that the model trained using SGLD can prevent the membership attack to a certain extent (formalized in Theorem~\ref{theo:uniform_bound}).

The most related work to ours was conducted by \citet{privacy_for_free_icml}, in which they introduced DP-SGLD that satisfies the differential privacy protocol~\cite{dwork2011differential}.
Our work differs from this work in two aspects. First, our theoretical analysis can be conducted without modifying the vanilla SGLD algorithm and
can be easily extended to other SG-MCMC methods. Second, differential privacy is a strict requirement for most real applications and leads
to a significant performance drop to meet the given privacy budget. DP-SGLD suffers such a performance drop~\cite{privacy_for_free_icml}. In contrast, we relax the requirement of differential privacy and focus on studying membership privacy (instead of differential privacy) afforded by SGLD.

Moreover, we build the theoretical
connection between DP-SGLD and our framework (see more details in the section of theoretical analysis). Our theoretical analysis can be further
naturally extended to other types of SG-MCMC methods, such as Stochastic Gradient Hamiltonian Monte Carlo~(SGHMC)~\cite{SGHMC}. 

To verify our theoretical findings, we perform membership attacks on different real-world datasets to evaluate
the information leakage of models trained with different optimization methods~(e.g. SGD and SGLD). Empirical results show that SGLD can 
significantly reduce the information leakage of the training dataset in contrast to SGD and its variants. Besides, we also observe that SGLD can prevent DNNs from overfitting in some cases, which implicitly verifies the connection between overfitting and information leakage~\cite{overfitting_privacy}.

In summary, our contributions are three folds:
\begin{itemize}
    \item We introduce a theoretical framework to analyze the membership information leakage of a well-trained DNN model.
    \item With the theoretical framework, we show that the model trained using SGLD can prevent the membership attack to a certain extent.
    \item 
    Our framework provides a uniform interpretation for some previous theoretical results by incorporating them into our framework.
\end{itemize}
The rest of this paper is organized as follows. We first provide a brief review of the related work. We then demonstrate the theoretical analysis, including the membership privacy analysis framework and our main theoretical results, followed by the the empirical results to validate our theoretical findings. Finally, we conclude our work in the last section. 

\section{Related Work}
\label{sec:related}
\nosection{Bayesian Deep Learning}
Avoiding overfitting of the deep learning model is a longstanding goal for the research community. Bayesian deep learning is recently regarded
as an intrinsic way to eliminate overfitting through ensembling models
sampled from the Bayesian posterior~(i.e. approximate Bayesian inference)~\cite{BBB}. As the Bayesian posterior is always intractable in practice, there are normally two different ways to approximate the
posterior sampling, i.e. Stochastic Variational Inference~(SVI)~\cite{BBB} and Stochastic Gradient
Markov Chain Monte Carlo~(SG-MCMC)~\cite{sgld_original,SGHMC}. The key idea of SVI based methods
is to use a parameterized variational distribution to approximate the real posterior. The approximation can be further transformed into an
optimization problem. To bring this idea into the realm of deep learning, previous works always come at the assumption that the Bayesian posterior of the parameters can be decomposed into separable
Gaussian distributions~\cite{BBB,psgld}. However, this assumption is always violated, which leads to non-negligible approximation error~\cite{psgld}. An alternative way is
to use SG-MCMC methods. In contrast to SVI based methods, the key idea of SG-MCMC is to approximately sample from the real posterior instead
of directly approximating the posterior~\cite{sgld_original}. There are various SG-MCMC methods, including Stochastic Gradient Langevin Dynamics~(SGLD) and Stochastic Gradient Hamiltonian Monte Carlo
~(SGHMC)~\cite{SGHMC}. In this paper, we focus on studying privacy related properties of SGLD, one of the most commonly used SG-MCMC methods. 

\nosection{SGLD}
The idea of combining stochastic optimization \cite{stochastic_optimization} and Langevin Dynamics \cite{langevin_dynamics} was
first presented in the work \cite{sgld_original}, in which the authors
proposed Stochastic Gradient Langevin Dynamics~(SGLD) to efficiently perform Bayesian learning on large-scale datasets. In contrast to SGD, 
SGLD injects proper Gaussian noise into the gradient estimation to avoid collapsing to just the maximum a posterior~(MAP) solution. 
Since then, many variants of SGLD are introduced to improve the vanilla SGLD algorithm. For example, \citet{psgld} proposed pSGLD, imposing a
preconditioner to the optimization process for improving the convergence efficacy of SGLD. Besides the algorithmic improvements of SGLD, there are a number of works focusing on applying SGLD to various applications \cite{sgld_shape_classification,sgld_language_model}. For instance, \citet{sgld_language_model} proposed using SGLD to mitigate the overfitting of RNN models and achieved
remarkable performances on several language modeling tasks. Among the above-mentioned works, there is another research direction to study the theoretical properties of SGLD.   
One of the most important properties of SGLD is that it can approximate the sampling from the posterior distribution~\cite{sde_view_sgld,higher_order}. This property has been theoretically studied in different
works. A popular way is to conduct the approximation analysis of SGLD from the perspective of stochastic differential equations. A recent research trend considers the generalization
bound of the SGLD algorithms under different assumptions~\cite{colt_sgld_gen,ISIT_sgld_gen}. In this paper, we take a totally different view, to study SGLD from the aspect of membership privacy. 

\nosection{Membership Attack}
The privacy leakage issue of machine learning models has attracted ever-rising interests from both the machine learning and security communities. 
Recently, membership attacks have arisen as a common threat model against machine learning algorithms \cite{stolen_memorize,s_and_p_attack,member_gen}. 
A pioneering work~\cite{s_and_p_attack} investigated the risk of membership attacks on different machine learning models. Specifically, they developed a shadow training technique to obtain an attack model in the black-box setting (i.e., without knowing the machine learning model structure and parameters). \citet{secret_sharer} proposed a metric to measure the vulnerability of deep learning models. A recent work \cite{optimal_attack_ICML} introduced the concept of the optimal attack and employed
the Bayes theorem to derive a formulation for the attack. In this paper, we take the membership attack as the workhorse for evaluating the information leakage of deep learning models trained
using SGLD.
\section{Theoretical Analysis}
\label{sec:theoretical_analysis}
In this section, we theoretically analyze the properties of the SGLD algorithm from different aspects that are related to privacy protection. We first introduce some basic notations and settings, including the SGLD algorithm and the membership attack. Then we build the theoretical framework to characterize the information leakage
of the trained model. Based on the framework, we demonstrate the main theoretical finding of this paper, i.e., the model trained using SGLD can preserve the membership privacy (preventing the membership attack). At last, we provide some other theoretical perspectives and show how to incorporate them into our framework, which helps us to
give a uniform understanding of the SGLD algorithm.
\subsection{Notation and Setting}
We start with basic notations and settings. In this paper, we focus on the classification task. Note that our analysis can be easily extended to other types of learning problems, such as regression.
In a supervised classification task, there is a labeled training dataset $\mathcal{D} = \{(x_i, y_i)\}_{i=1}^{N}$, where
$x$ denotes the input data (e.g. an image in the image classification task) and $y$ denotes
the associated label. For simplicity, we denote $z_i=(x_i, y_i)$ as a data-label pair or a sample. 
The goal of a deep learning model $\mathbb{M}$ (e.g. a DNN model) is to assign
a probability to each possible label $y$, i.e. $\mathbb{M}$ is a
probabilistic model that computes $p(y|x;\boldsymbol{\theta}) = f(x, \boldsymbol{\theta})$, where 
$\boldsymbol{\theta}$ denotes the weight parameters of $\mathbb{M}$ and $f$ denotes the inference of $\mathbb{M}$. A typical loss used for optimizing
$\boldsymbol{\theta}$ is the negative log-likelihood which
satisfies the following equation:
\begin{equation}
    L(\mathcal{D}, \boldsymbol{\theta}) = \sum_{z_i\in \mathcal{D}} l(z_i, \boldsymbol{\theta})\text{,}
    \label{eq:ld_loss}
\end{equation}
where $l(z, \boldsymbol{\theta})$ is the negative log-likelihood over the individual sample $z = (x,y)$,
i.e. $l(z,\boldsymbol{\theta}) = -\log(p(y|x;\boldsymbol{\theta}))$.

\nosection{SGLD}
SGLD can be regarded as an effective optimization method
that incorporates uncertainty into the weight parameter.
Based on the above notations, we can define the
updating rule of SGLD as:
\begin{equation}
    \boldsymbol{\theta}_{t+1} = \boldsymbol{\theta}_{t}-(\dfrac{\epsilon_t}{2}\partial_{\boldsymbol{\theta}}L(\mathcal{B}_t, \boldsymbol{\theta}_t)+\eta_t), \eta_t\sim\mathcal{N}(0, \epsilon_t \mathbf{I})\text{,}
    \label{eq:sgld}
\end{equation}
\begin{equation}
     L(\mathcal{B}_t, \boldsymbol{\theta}_t)= -\log p(\boldsymbol{\theta}_t)+\dfrac{|\mathcal{D}|}{|B_t|}\sum_{ z_i \in \mathcal{B}_t}l(z_i, \boldsymbol{\theta}_t)\text{,}
\label{eq:sgld_loss}
\end{equation}
where $p(\boldsymbol{\theta}_t)$ is the prior of $\boldsymbol{\theta}_t$ and $\epsilon_t$ is the step size. $\mathcal{N}(0, \epsilon_t\mathbf{I})$ denotes the Gaussian distribution. $L(\mathcal{B}_t,\boldsymbol{\theta}_t)$ can be seen as
a stochastic approximation of Equation~\eqref{eq:ld_loss}, which is to compute the negative log-likelihood over
the mini-batch $\mathcal{B}_t$ that is randomly selected
from the whole dataset $\mathcal{D}$.

Previous studies always assume the step size decreases towards zero at the rate that satisfies some conditions~\cite{sgld_original}.
Due to this assumption, intuitively, the optimization process
of SGLD can be divided into two phases in terms of the
step size. In the initial phase, the stochastic gradient noise\footnote{The noise is caused by the random sampling.} will dominate the optimization process due to the
large step size. The SGLD algorithm in this phase can be seen as an efficient stochastic gradient descent algorithm. The initial phase is also called as the burn-in phase in the literature~\cite{sgld_original}. In the later phase, the injected Gaussian noise will dominate the process
as the step size gets smaller. The parameter obtained by the SGLD algorithm in this phase can be
seen as sampling from the true posterior distribution. 
More rigorous descriptions can be found in prior works~\cite{sgld_original}. 

\nosection{Membership Attack} 
The membership attack refers to inferring whether a specific
sample is used for training the model.
Formally, given a specific sample $z = (x, y)$ and a DNN model $\mathbb{M}$ with the parameter $\boldsymbol{\theta}$, the attacker aims to compute the probability that
the sample belongs to the training data. Here we denote the probability
as $P(m=1|z,\boldsymbol{\theta})$, where $m$ refers to the sample's membership and $m=1$ indicates that the sample is in the training dataset. Besides the knowledge of the trained model,
the attacker may also hold an extra dataset $\mathcal{S}$, which consists of samples from the training/hold-out datasets.
We call this dataset ``shadow dataset'' borrowed from the prior work\footnote{This can also be seen as the side information in the literature.}. We denote this dataset as $\mathcal{S}=\{(z_i, m_i)\}_{i=1}^n$, where each $z_i$ is a sample of $(x_i, y_i)$ and $m_i$ represents the
membership of this sample. The samples with $m_i=1$ are from the training dataset, otherwise, are from the hold-out dataset. Based on the above setting, the
attacker can employ the shadow dataset $\mathcal{S}$ to
assist the membership inference, which is to compute the probability $P(m=1|z,\boldsymbol{\theta}, \mathcal{S})$. Based on
the previous work~\cite{optimal_attack_ICML}, we call the exact value of $P(m=1|z,\boldsymbol{\theta}, \mathcal{S})$ as the optimal attack. In practice, the optimal attack
is always intractable thus previous works have proposed different methods to approximate $P(m=1|z,\boldsymbol{\theta}, \mathcal{S})$~\cite{optimal_attack_ICML}. In what follows, we introduce
our theoretical framework to analyze the information leakage~(in the aspect of membership privacy) of the learned deep learning models.

\subsection{Membership Privacy Analytical Framework}
We further assume the ratio of training samples\footnote{Training samples are those from the training dataset.} in $\mathcal{S}$ is $\lambda$. Thus we can give a trivial
attack solution that proceeds as $P(m=1|z,\boldsymbol{\theta}, \mathcal{S})\approx P(m=1)=\lambda$. In contrast to the optimal attack mentioned above, we call this attack as the trivial attack, since this attack gets nothing other than the macroscopical information of the shadow dataset. We can define the membership
privacy leakage~(Mpl) loss of a given sample as:
\begin{equation}
    {\rm Mpl}(z, \theta, \mathcal{S})= \max(P(m=1|z,\boldsymbol{\theta}, \mathcal{S})-\lambda,0)\text{.}
    \label{eq:mpl}
\end{equation}
In the above equation, when $P(m=1|z,\boldsymbol{\theta},\mathcal{S})\leq \lambda$, the optimal attack
infers that the sample $z$ is not from the training dataset. Thus we set the Mpl loss to zero since the optimal attack do not induce the membership leakage of the training dataset.
The expectation of the Mpl loss over all possible samples~(i.e. $\mathbb{E}_z [{\rm Mpl}(z, \boldsymbol{\theta}, \mathcal{S})]$) can evaluate the difference
between the optimal attack and the trivial attack. Obviously, the smaller the
expectation of the Mpl loss is, 
the less information of the training dataset is revealed by the model~(i.e. the optimal attack is more similar to the trivial attack).
To compute the loss, we need to give an implicit formulation of the optimal attack. In this paper, we introduce a modified version of the formulation proposed by~\citet{optimal_attack_ICML}.
To this end, we first define the posterior distribution over the parameter~(while $\mathcal{S}$ is given) by:
\begin{equation}
    p_{\mathcal{S}}(\boldsymbol{\theta})= p(\boldsymbol{\theta}|\mathcal{S}_T), \mathcal{S}_T=\{(z_i, m_i)\in \mathcal{S}: m_i=1\}.
\label{eq:ps}
\end{equation}
With a slight modification of the prior work~\cite{optimal_attack_ICML}, we can obtain: 
\begin{lemma}
    \cite{optimal_attack_ICML} Given a learned model with parameter $\boldsymbol{\theta}$, the shadow dataset
    $\mathcal{S}$, and a specific sample $\mathbf{z}$, the Mpl loss can be computed using:
\begin{equation}
P(m=1|z, \boldsymbol{\theta}, \mathcal{S}) = 
\sigma(s(z, \boldsymbol{\theta}, p_{\mathcal{S}})+t_\lambda)\text{,}
\end{equation}
\label{lemma:bound1}
\end{lemma}
where $t_{\lambda}$ is {\small $\log(\dfrac{\lambda}{1-\lambda})$} and $\sigma(t):=1/(1+e^{-t})$ is the sigmoid function. We call $s$ the \emph{measure score} which satisfies the following equations:
\begin{equation*}
    t_p(z) = -\log\mathbb{E}_{\boldsymbol{\theta}\sim p}e^{-l(z, \boldsymbol{\theta})}\text{,}
    \label{eq:threshold}
\end{equation*}
\begin{equation*}
    s(z, \boldsymbol{\theta}, p) = t_p(z)-l(z, \boldsymbol{\theta})\text{.}
    \label{eq:meaure_function}
\end{equation*}
We can further use the {Lipschitz continuity of  $\sigma$ that $|\sigma(u)-\sigma(v)| \leq |u-v|/4$ and $\lambda = \sigma(t_{\lambda})$} to bound the Mpl loss by:
\begin{equation}
    {\rm Mpl}(z, \boldsymbol{\theta}, \mathcal{S}) \leq
{\max(s(z, \boldsymbol{\theta}, p_{\mathcal{S}})/4,0)}\text{.}
\label{eq:mpl_bound1}
\end{equation}
The proof of Equation~\eqref{eq:mpl_bound1} can be found in the appendix.
Equation~\eqref{eq:mpl_bound1} points out that the Mpl loss
can be bounded by the measure score.
Intuitively, the value of $t_p(z)$ can be seen as a threshold which is used for comparing with the loss function with 
respect to $z$ and the trained parameter $\boldsymbol{\theta}$. From Equation~\eqref{eq:mpl_bound1}, if the negative log-likelihood of the given sample $z$ is close to the threshold, i.e., $l(z, \boldsymbol{\theta})\approx t_p(z)$, then the measure score is close to zero, thus the optimal attack can not obtain any information
other than the prior knowledge (i.e., $P(m=1)=\lambda$). And when the measure score is substantially greater/less than zero~(i.e., the loss of the given sample is much lower/higher than the threshold), the optimal attack can gain non-trivial membership information on $z$, i.e., $P(m=1|z, \boldsymbol{\theta})\gg \lambda$. 

To enable the analysis on the SGLD algorithm, we need to consider the case of model ensemble used for approximating Bayesian inference. We further reformulate the above Equation~\eqref{eq:mpl_bound1} based on two facts. First, we can use Jensen's inequality
to obtain:
\begin{equation}
    t_p(z) \leq \mathbb{E}_{\boldsymbol{\theta}\sim p}[l(z,\boldsymbol{\theta})]\text{.}
    \label{eq:s_bound}
\end{equation}
Second, we note that, when the sample $z$ and parameter $\boldsymbol{\theta}$ are given, the value of $l(z,\boldsymbol{\theta})$ explicitly depends on the
predicted probability with respect to the label $y$. Here, we denote the predicted probability as $\hat{y} = f(x, \boldsymbol{\theta})$. With a slight notation
abuse, we rewrite the value of $l(z, \boldsymbol{\theta})$ as $\tilde{l}(y, \hat{y})$. In the case of SGLD ensemble, $\hat{y}$ can be obtained by average outputs of models produced by
the SGLD algorithm. 
Then, combining with Equation~\eqref{eq:s_bound} and slightly modifying the above derivation~(i.e. replacing $\boldsymbol{\theta}$ with $\hat{y}$ ), we can reformulate the inequality in Equation~\eqref{eq:mpl_bound1} as:
\begin{equation}
    {\rm \widetilde{Mpl}}(z, \hat{y}, \mathcal{S}) \leq \max(\tilde{s}(z, \hat{y}, p_{S})/4,0)\text{,}
    \label{eq:mpl_bound2}
\end{equation}
where:
\begin{equation}
\tilde{s}(z, \hat{y}, p_{S}) = \mathbb{E}_{\boldsymbol{\theta}\sim p_{\mathcal{S}}}[l(z,\boldsymbol{\theta})]-\tilde{l}(y,\hat{y})\text{,}
\label{eq:s_hat}
\end{equation}
where $\hat{y}$ is the average output of models sampled using SGLD. In contrast to Equation~\eqref{eq:mpl_bound1} that can only be applied
to the context of the single model, we can leverage Equation~\eqref{eq:mpl_bound2} to analyze the information leakage of the ensembled models\footnote{Because Equation ~\eqref{eq:mpl_bound2} only depends on the output vector.}. This is appealing for the analysis of SGLD, since its key feature is to approximate the Bayesian inference via ensembling the models produced by SGLD after the burn-in phase~\cite{sgld_original}.   
\subsection{Uniform Bound for the Mpl Loss}
Up to now, we can leverage the above analytical framework to quantify the information leakage of the trained model in terms of the membership privacy loss. In this part,
for the model trained using SGLD, we aim to give a uniform bound of the Mpl loss for arbitrary sample in the worst case. Here, the worst case refers to the shadow dataset $\mathcal{S}$ held by the attack is a leave-one-out version of the original training dataset $\mathcal{D}$, i.e., $S$
is obtained by removing one sample from $\mathcal{D}$. Without loss of generality, we assume the removed sample is $z_n$, i.e.
$\mathcal{S}=\mathcal{D}/z_n$. Here, we assume that the samples in $\mathcal{S}$ are all from the training dataset, since the posterior $p_\mathcal{S}$ (Equation~\eqref{eq:ps})
does not depend on the samples from the hold-out set. The analysis can be easily extended to other scenarios (e.g., $\mathcal{S}$ has samples from the hold-out dataset).
To introduce the following results, we first give a commonly used assumption presented in the prior work~\cite{privacy_for_free_icml}.
\begin{assumption}
We assume $z$ is sampled from the space $\mathcal{Z}$ and $\boldsymbol{\theta}$ is sampled from the SGLD algorithm. This assumption limits the negative log-likelihood to be bounded, i.e., $\sup_{z\in\mathcal{Z},\boldsymbol{\theta}}|l(z,\boldsymbol{\theta)}|\leq B$. 
The step sizes $\{\epsilon_t\}$ in Equation~\eqref{eq:sgld} decrease and satisfy:
1) $\sum_{t=1}^{\infty}\epsilon_t=\infty$, and 2) $\lim_{L\to\infty}\dfrac{\sum_{t=1}^{L}\epsilon_t^2}{S_t}=0$, where $S_t=\sum_{t=1}^{L}\epsilon_t$ is the sum of the step sizes. 
\end{assumption}
Formally, based on the above settings, we introduce the main theoretical finding as the following theorem:
\begin{theorem}(\textbf{Uniform bound in expectation})
Given $\mathcal{S}=\mathcal{D}/z_n$ and model parameters $\{\boldsymbol{\theta}_t\}_{t=t_0}^{t_0+L}$ generated by SGLD after the burn-in phase that starts from the $t_0$-th iteration.
We denote the sampling process as ${\boldsymbol{\theta}\sim SGLD}$. $\hat{y}$ is the weighted average output using these parameters, i.e., $\hat{y}=\sum_{t=t_0}^{t_0+L}\dfrac{\epsilon_t}{S_L}f(x,\boldsymbol{\theta}_t)$, where $S_L=\sum_{t=t_0}^{t_0+L}\epsilon_t$ . Under Assumption 1, we can bound the expectation of the {\rm Mpl} loss
for any sample $z$ by a uniform constant $M_{SGLD}$:
\begin{equation}
  \mathbb{E}_{SGLD}[{\rm \widetilde{Mpl}}(z,\hat{y}, \mathcal{S})] \leq M_{SGLD}\text{,}
\end{equation}
\begin{equation}
    M_{SGLD} = \frac{B}{4}(e^{2B}-1)+O(\dfrac{1}{S_L}+\dfrac{\sum_{t=t_0}^{t_0+L}\epsilon_t^2}{S_L})\text{.}
\end{equation}
\label{theo:uniform_bound}
\end{theorem}
The expectation is with respect to the randomness of SGLD (i.e., sampling different parameters to compute the predicted value $\hat{y}$).
Here, for simplicity, Theorem~\ref{theo:uniform_bound} is for the case of decreasing-step-size~\cite{sgld_original} and we can easily extend this theorem to the case of constant-step-size
based on the prior work~\cite{sde_view_sgld}.
To given a proof of Theorem~\ref{theo:uniform_bound}, we first introduce two lemmas. 
\begin{lemma}
    \textbf{(Bound of the leave-one-out error)}~\cite{colt_sgld_gen} Given $\mathcal{S}=\mathcal{D}/z_n$, under Assumption 1, we can bound the leave-one-out error (i.e. the posterior average difference with respect to these two datasets) as:
    \begin{equation}
        |\mathbb{E}_{\boldsymbol{\theta}\sim p_{\mathcal{S}}}[l(z,\boldsymbol{\theta})]-\mathbb{E}_{\boldsymbol{\theta}\sim p_{\mathcal{D}}}[l(z, \boldsymbol{\theta})]| \leq B(e^{2B}-1)\text{.}
    \end{equation}
    \label{lemma:loo-bound}
\end{lemma}
The proof of Lemma~\ref{lemma:loo-bound} can be found in the appendix.

The following lemma is based on previous finite-time error analysis of SGLD~\cite{higher_order}. 
\begin{lemma}
    \textbf{(Finite-time error analysis)}~\cite{higher_order} Given model parameters $\{\boldsymbol{\theta}_t\}_{t=t_0}^L \sim SGLD$ generated by SGLD after the burn-in phase and the predicted value $\hat{l}= \sum_{t=t_0}^{t_0+L}\dfrac{\epsilon_t}{S_L}l(z,\boldsymbol{\theta}_t)$. $S_L=\sum_{t=t_0}^{t_0+L}\epsilon_t$ is the sum of the step sizes. Then we have:
    \begin{equation}
        |\mathbb{E}_{SGLD}[\hat{l}]-\mathbb{E}_{\boldsymbol{\theta} \sim p_{\mathcal{D}}}[l(z, \boldsymbol{\theta})]|= O(\dfrac{1}{S_L}+\dfrac{\sum_{t=t_0}^{t_0+L}\epsilon_t^2}{S_L})\text{.}
    \end{equation}
    \label{lemma:approx_error}
\end{lemma}
Here, with the help of the above two lemmas, we demonstrate the roadmap to accomplish the proof of Theorem~\ref{theo:uniform_bound}. The details of the proof can be found in the appendix.
\begin{proof}
\textbf{(Sketch proof of Theorem~\ref{theo:uniform_bound})}: 
{\rm Based on Equation~\eqref{eq:mpl_bound2}, we can first bound the measure score, which can be further
used for bounding the Mpl loss. 

To bound the measure score, we can decompose Equation~\eqref{eq:s_hat} into the leave-one-out error and the approximation error:
\begin{equation*}
\begin{split}
    \tilde{s}(z, \hat{y}, p_{S}) &= \underbrace{\mathbb{E}_{\boldsymbol{\theta}\sim p_{\mathcal{S}}}[l(z,\boldsymbol{\theta})]-\mathbb{E}_{\boldsymbol{\theta}\sim p_{\mathcal{D}}}[l(z, \boldsymbol{\theta})]}_{leave-one-out~error}\\
    &+\underbrace{\mathbb{E}_{\boldsymbol{\theta}\sim p_{\mathcal{D}}}[l(z, \boldsymbol{\theta})]-\tilde{l}(y,\hat{y})}_{approximation~error}\text{.}
\end{split}
\end{equation*}
Then we can obtain the bounds for these two errors individually. First, we notice that the leave-one-out error can be bounded based on
    Lemma \ref{lemma:bound1}. 
    Then the approximation error can be further decomposed into:
    \begin{equation*}
        \underbrace{\mathbb{E}_{\boldsymbol{\theta}\sim p_{\mathcal{D}}}[l(z, \boldsymbol{\theta})]- \hat{l}}_{Factor 1}+\underbrace{\hat{l}-\tilde{l}(y,\hat{y})}_{Factor 2}\text{.}
    \end{equation*}
    Note that factor 1 can be bounded via Lemma~2 when taking the expectation over parameters generated by SGLD (the boundedness is ensured by Assumption 1 and Lemma~\ref{lemma:approx_error} jointly). 
    
    The boundedness of factor 2 is ensured by the boundedness
    of the negative log-likelihood and the boundedness of the variance of SGLD (see details in the appendix).
    
    Therefore we can obtain the bound of the measure score in the context of SGLD model
    ensemble.}
\end{proof}
\subsection{Other Theoretical Perspectives}
In this part, we revisit two theoretical results provided in previous works, the generalization of SGLD~\cite{colt_sgld_gen} and the DP-SGLD~\cite{privacy_for_free_icml}. We will show how to give a unified interpretation of these results from the aspect of membership privacy. 

\nosection{Connection to Differential Privacy}
Previously, \citet{privacy_for_free_icml} presented DP-SGLD, which modifies the existing SGLD algorithm to satisfy differential privacy~\cite{dwork2011differential}. The key idea of DP-SGLD is to
calibrate the magnitude of the injected Gaussian noise and use the strong composition theory to obtain the privacy budget. Here, we show that DP-SGLD can also
preserve membership privacy in terms of the boundedness of the Mpl loss. 
Intuitively, since DP-SGLD satisfies $(\epsilon, \delta)$-differential privacy, each
sample in the training dataset will have a limited impact on the learned model. Therefore,
it is more difficult to detect useful membership information of an individual sample.
This intuition can be theoretically interpreted using the following property: 
\begin{property}
(\textbf{DP-SGLD}) Given $\boldsymbol{\theta}$ learned using $(\epsilon, \delta)$-DP-SGLD, 
then the following inequality holds with probability $1-\delta$:
\begin{equation}
    {\rm Mpl}(z, \boldsymbol{\theta}, \mathcal{S}) \leq \dfrac{\epsilon}{4}\text{.}
\end{equation}
\end{property}
The proof of this property can be found in the appendix.
Here we only consider the case
of a single model without model ensemble.
 This theorem can also be extended to
the case of model ensemble by using the composition theory of differential privacy. 

\nosection{A View From Generalization}
 Previous works have validated the connection between the overfitting and the information leakage of DNN
 models~\cite{wu2019p3sgd,overfitting_privacy}. In the realm of membership privacy, \citet{member_gen} also
 pointed out that overfitting is a sufficient condition for exposing  membership information of the training 
dataset. This finding inspires us to understand why SGLD works for preventing the membership attack from
the view of the generalization theory. Specifically, based on the prior work~\cite{colt_sgld_gen},
we can prove that SGLD can provide the property of uniform stability. Armed with the stability learning theory~\cite{shwartz_stable}, \citet{colt_sgld_gen} further provided an expectation bound for the generalization error of  
the SGLD algorithm. Since the generalization error can be bounded, it is challenging for the attacker to notice the output's (w.r.t. the model) difference between the training and hold-out datasets, which further prevents the membership
attack (e.g., the threshold attack conducted in our experiments). Therefore, from the view of the generalization bound, we can understand the effectiveness of SGLD to prevent the membership attack. 
\section{Evaluation}
\subsection{Experimental Settings}
\label{sec:exp}

\nosection{Datasets}
In this paper, we choose datasets from scenarios where data privacy is important, such as financial and medical data analysis. Specifically, we select two datasets as our benchmarks, namely, German Credit dataset~\cite{uci_dataset} and IDC dataset\footnote{{\tt http://www.andrewjanowczyk.com/use-case-6-}\\{\tt invasive-ductal-carcinoma-idc-segmentation/}}. 

The German Credit dataset consists of $1,000$ applications for credit cards. Each application is labeled with good or bad credit. We consider the classification task to identify ``good credit" applications. We randomly split the whole dataset
into training ($400$ applications), hold-out/validation ($300$ applications), and test~($300$ applications) sets. Here, the hold-out dataset can be used for either validation or building the ``shadow'' attack model~\cite{s_and_p_attack}.  

IDC dataset is used for 
invasive ductal carcinoma (IDC) classification. This dataset contains $277\rm{,}524$ patches of $50\times50$ pixels ($198\rm{,}738$ IDC-negative and $78\rm{,}786$ IDC-positive). Following the setting of the work \cite{stolen_memorize},
we split the whole dataset into training, validation~(hold-out), and test sets. To be specific, the training dataset consists of $10,788$ positive patches and $29,164$ negative patches. The test dataset consists of 
$11,595$ positive patches and $31,825$ negative patches. The remain patches are used as the hold-out dataset.

\nosection{Model Setup} 
For the German Credit dataset~\cite{uci_dataset}, we choose a three-layer fully-connected neural network as the model architecture.
For comparison, we train the model using both SGLD and SGD variants. All these training strategies share the following hyper-parameters: the mini-batch is set to 32 and the epoch number is set to 30. The learning rate decreases by half every 5 epochs. For SGLD, the variance of the prior is set to $1.0$. The initial learning rate is set to $1\times10^{-3}$. 

For the IDC dataset, since the training dataset comprises image data, we choose a typical
residual convolutional neural network, ResNet-18~\cite{he2016}. The mini-batch
is set to 128 and the epoch number is set to 100. Data augmentation is not used. The learning rate decreases by half every 20 epochs.
For SGLD, the variance $\sigma^2$  of the prior is set to 1.0. The initial learning rate is set to $1\times10^{-4}$.

\nosection{Attack Setup}
The key part of our experiments is to quantitatively evaluate the information leakage of the model trained using different methods. In practice, we found that a straightforward
attack, \emph{threshold attack} \cite{overfitting_privacy}, suffices to support our theoretical findings. The threshold attack has already been used in various previous works \cite{overfitting_privacy}. This attack is based on the
 intuition, i.e., a sample with lower loss is more likely to belong to the training dataset. We denote $\mathcal{A}$ as the attacker and the threshold attack proceeds as follows (here, we use the consistent notations as in
the theoretical analysis):
\begin{itemize}
    \item Given a specific sample $z$ and its predicted probability $\hat{y}$ computed by the model.
    \item $\mathcal{A}$ calculates the normalized loss $n=\tilde{l}(y, \hat{y})/B$, where $B$ is the bound for the negative log-likelihood.
    \item $\mathcal{A}$ selects a threshold $t$ and outputs $1$ (indicates the sample is a training sample) if $n\leq t$. Otherwise, it outputs $0$.
\end{itemize}
The threshold attack is evaluated on the training dataset and the test dataset. Note that we can choose different threshold $t$ to obtain different attack models. Naturally,
we can plot the ROC curve of these attack results and use the area under the ROC curve (AUC) as an indicator of the information leakage. Moreover, we assume that the
attacker can obtain the average value of $\tilde{l}(y, \hat{y})$ on the training dataset. Thus we can directly use this value as a threshold to build the attack model, and compute the F1 score and accuracy of the attack model to quantify the amount of information leakage. In summary, we use three metrics to evaluate the information leakage of a learned model, namely, the AUC value, the F1 score, and the accuracy. 

\subsection{Experimental Results}
In Table~\ref{tab:gcd} and \ref{tab:idc}, Dropout denotes
adding Dropout layers into the network architecture. For ResNet-18, we insert Dropout between convolutional layers and set the drop ratio to $0.3$ following the work~\cite{wider-resnet}.
SGLD+ represents ensembling models obtained by SGLD. In practice, we found that ensembling 3 models
suffices to our case, thus we use the models obtained in the last three iterations and average their
outputs for the final prediction. In contrast to the decreasing step size discussed in the theoretical analysis, we use the constant step size in practice following prior works~\cite{sde_view_sgld}. For a fair comparison with SGLD+,
we also apply the same ensemble strategy to the case of SGD (denoted as SGD+).

\nosection{German Credit Dataset}
The overall results of the German Credit dataset are shown in Table~\ref{tab:gcd}. We observe
that the use of SGLD can prevent the information leakage of the training dataset, based on the above attack metrics for evaluating the information leakage. In contrast to SGD, SGLD significantly reduces the AUC score of the threshold attack
from 0.677 to 0.536, while achieves comparable model accuracy. We can also
observe consistent performance boost and information leakage reduction when model ensemble is used. For example,
SGLD+ decreases the F1 score of the threshold attack (from 0.598 to 0.564) while achieves a slight model accuracy
increase (from 0.736 to 0.751). 

\nosection{IDC Dataset}
The overall results are shown in Table~\ref{tab:idc}. SGLD can significantly reduce the information leakage in contrast to SGD, in terms of
the aforementioned metrics for evaluating the attack performance. For example, compared with SGD, SGLD remarkably reduces the AUC value of threshold attack from 0.669 to 0.641 while
achieving comparable model accuracy. We also observe a slight model accuracy improvement when the ensemble strategy is used (see the results of SGLD+). It is worth noting that Dropout is not effective in reducing the information leakage and improving the performance in this case (see more discussions in the following subsection).

All these empirical results validate our theoretical findings, i.e., the model trained using SGLD can preserve membership privacy to a certain extent.
\begin{table}
\centering
\caption{Results of the threshold attack on the German Credit dataset. SGLD+ represents model ensemble.}
\label{tab:gcd}
\begin{tabular}{ccccccc}
\toprule
\multirow{2}{*}{Strategy} & \multicolumn{3}{c}{Attack Metric} & \multicolumn{3}{c}{Model Metric} \\
                          & AUC         & F1         & Acc         & Train        & Test       & Gap   
                    \\
                    \midrule
SGD                       &0.677             &0.685            &0.646             &1.000              &0.734           &0.266           \\
SGD+                &0.677                       &0.684              &0.645                     &1.000              &0.720    &0.280\\
Dropout               &0.527             &0.567            &0.527             &0.859              &0.747            &0.112           \\
RMSprop                   &0.589             &0.677            &0.627             &0.965              &0.749            &0.216           \\ \midrule
SGLD                      &0.536             &0.598            &0.539             &0.811              &0.736           &0.075           \\
SGLD+             &0.536             &0.564            &0.526             &0.839              &0.751            &0.087           \\
pSGLD                     &0.551             &0.600            &0.550             &0.858              &0.753            &0.105          \\
pSGLD+           &0.550             &0.568            &0.539             &0.859              &0.758           &0.101          \\
\bottomrule
\end{tabular}
\end{table}
\begin{table}
\caption{Results of the threshold attack on the IDC dataset. SGLD+ represents model ensemble.}
\label{tab:idc}
\centering
\begin{tabular}{ccccccc}
\toprule
\multirow{2}{*}{Strategy} & \multicolumn{3}{c}{Attack Metric} & \multicolumn{3}{c}{Model Metric} \\
                          & AUC         & F1         & Acc         & Train        & Test       & Gap   
                    \\
                    \midrule
SGD                       &0.669             &0.716            &0.646             &0.993              &0.823            &0.170           \\
SGD+                  &0.665             &0.714            &0.638             &0.998              &0.837            &0.161 \\
Dropout               &0.662             &0.711            &0.634             &0.994              &0.818           &0.176           \\
RMSprop                   & 0.674            &0.715            &0.652             &0.992              &0.818            &0.174           \\ \midrule
SGLD                      &0.641             &0.669            &0.620             &0.973              &0.817            &0.156        \\
SGLD+             &0.643             & 0.658           &0.617             &0.981              &0.825           &0.156           \\
pSGLD                     &0.652             &0.697            &0.641             &0.982              &0.824            &0.158           \\
pSGLD+           &0.648             &0.692            &0.640             &0.987              &0.830            &0.157      \\
\bottomrule
\end{tabular}
\end{table}

\subsection{Discussions}

\nosection{SGLD Variants}
The vanilla SGLD algorithm updates all parameters with the same step size, which always leads to inefficient optimization since these parameters have different curvature. pSGLD~\cite{psgld} solves this problem by applying the preconditioner to each parameter in the optimization process. In this paper, we also demonstrate some
numerical results of pSGLD. Since pSGLD can be seen as a noisy version of the RMSprop algorithm, we also
show the results of RMSprop. From Table~\ref{tab:gcd}, we observe that pSGLD can improve the
test accuracy in contrast to the vanilla SGLD. We infer that the accuracy boost comes from the use of the preconditioner which can improve the optimization efficiency.
However, the accuracy boost accompanies with
more information leakage in terms of the above attack metrics. For example, comparing with SGLD, pSGLD increases the AUC value of the attack model
from 0.536 to 0.551. 
From the updating rule of pSGLD~\cite{psgld}, pSGLD uses more information of the gradient compared with SGLD, which causes the increase of the information leakage. 

\nosection{Effectiveness of Dropout}
As shown in Table~\ref{tab:gcd} and \ref{tab:idc}, Dropout demonstrates different effects on these two tasks since two different types of model architectures are used. According to previous works, Dropout is less effective on the fully convolutional neural network in which the fully-connected layer only exists in the last layer.~\cite{drop_block}. This conclusion coincides with our observation, i.e., the Dropout cannot boost the performance on the ResNet-18 model
for the IDC classification. On the contrary, in the case of the German Credit dataset, one notable observation is that Dropout
can also drastically reduce the information leakage~(see Dropout in Table~\ref{tab:gcd}). This result is not surprising to us, since Dropout can be seen as an implicit
Bayesian inference based on the prior work~\cite{Bayesian_dropout}. This suggests that we can use Dropout as an effective way to preserve the membership privacy for fully-connected networks (i.e. multilayer perceptron).

\nosection{Other Attacks}
In the above analysis, we found that the simple threshold attack suffices to detect some useful membership information in our case.  We also conduct experiments on other membership attack methods, such as shadow training method
proposed in the prior work \cite{s_and_p_attack}. And we found that SGLD is also effective to prevent such an attack in practice.

\section{Conclusions}
\label{sec:conclusion}
In this paper, we provided a novel perspective to study the properties of SGLD from the membership privacy. To this end, we first built the theoretical framework to
analyze the information leakage of a learned model. We then employed this framework to prove that the model learned using SGLD can prevent the membership attack
to a certain extent. Experiments on different real-world datasets have verified our theoretical findings. Moreover, our paper shed light on an interesting perspective of membership privacy
to explain behaviors of various deep learning algorithms, which indicates a new direction for the explainable deep learning research. 
\bibliographystyle{aaai}
\fontsize{9.6pt}{10.6pt} \selectfont
\bibliography{ref}

\begin{thebibliography}{}

\bibitem[\protect\citeauthoryear{Blundell \bgroup et al\mbox.\egroup
  }{2015}]{BBB}
Blundell, C.; Cornebise, J.; Kavukcuoglu, K.; and Wierstra, D.
\newblock 2015.
\newblock Weight uncertainty in neural networks.
\newblock {\em CoRR} abs/1505.05424.

\bibitem[\protect\citeauthoryear{Carlini \bgroup et al\mbox.\egroup
  }{2018}]{secret_sharer}
Carlini, N.; Liu, C.; Kos, J.; Úlfar Erlingsson; and Song, D.
\newblock 2018.
\newblock The secret sharer: Measuring unintended neural network memorization
  and extracting secrets.
\newblock {\em {ArXiv e-prints}} 1802.08232.

\bibitem[\protect\citeauthoryear{Chen, Ding, and Carin}{2015}]{higher_order}
Chen, C.; Ding, N.; and Carin, L.
\newblock 2015.
\newblock On the convergence of stochastic gradient {MCMC} algorithms with
  high-order integrators.
\newblock In {\em Advances in Neural Information Processing Systems 28:
  December 7-12, 2015, Montreal, Quebec, Canada},  2278--2286.

\bibitem[\protect\citeauthoryear{Chen, Fox, and Guestrin}{2014}]{SGHMC}
Chen, T.; Fox, E.~B.; and Guestrin, C.
\newblock 2014.
\newblock Stochastic gradient hamiltonian monte carlo.
\newblock In {\em Proceedings of the 31th International Conference on Machine
  Learning, {ICML} 2014, Beijing, China, 21-26 June 2014},  1683--1691.

\bibitem[\protect\citeauthoryear{Dua and Graff}{2017}]{uci_dataset}
Dua, D., and Graff, C.
\newblock 2017.
\newblock {UCI} machine learning repository.

\bibitem[\protect\citeauthoryear{Dwork}{2011}]{dwork2011differential}
Dwork, C.
\newblock 2011.
\newblock Differential privacy.
\newblock {\em Encyclopedia of Cryptography and Security}  338--340.

\bibitem[\protect\citeauthoryear{Fredrikson, Jha, and
  Ristenpart}{2015}]{ccs_model_inversion}
Fredrikson, M.; Jha, S.; and Ristenpart, T.
\newblock 2015.
\newblock Model inversion attacks that exploit confidence information and basic
  countermeasures.
\newblock In {\em Proceedings of the 22nd {ACM} {SIGSAC} Conference on Computer
  and Communications Security, Denver, CO, USA, October 12-16, 2015},
  1322--1333.

\bibitem[\protect\citeauthoryear{Gal and Ghahramani}{2016}]{Bayesian_dropout}
Gal, Y., and Ghahramani, Z.
\newblock 2016.
\newblock Dropout as a bayesian approximation: Representing model uncertainty
  in deep learning.
\newblock In {\em Proceedings of the 33nd International Conference on Machine
  Learning, {ICML} 2016, New York City, NY, USA, June 19-24, 2016},
  1050--1059.

\bibitem[\protect\citeauthoryear{Gan \bgroup et al\mbox.\egroup
  }{2017}]{sgld_language_model}
Gan, Z.; Li, C.; Chen, C.; Pu, Y.; Su, Q.; and Carin, L.
\newblock 2017.
\newblock Scalable bayesian learning of recurrent neural networks for language
  modeling.
\newblock In {\em Proceedings of the 55th Annual Meeting of the Association for
  Computational Linguistics, {ACL} 2017, Vancouver, Canada, July 30 - August 4,
  Volume 1: Long Papers},  321--331.

\bibitem[\protect\citeauthoryear{Ghiasi, Lin, and Le}{2018}]{drop_block}
Ghiasi, G.; Lin, T.; and Le, Q.~V.
\newblock 2018.
\newblock Dropblock: {A} regularization method for convolutional networks.
\newblock In {\em Advances in Neural Information Processing Systems 31: NeurIPS
  2018, 3-8 December 2018, Montr{\'{e}}al, Canada.},  10750--10760.

\bibitem[\protect\citeauthoryear{He \bgroup et al\mbox.\egroup }{2016}]{he2016}
He, K.; Zhang, X.; Ren, S.; and Sun, J.
\newblock 2016.
\newblock Deep residual learning for image recognition.
\newblock In {\em 2016 {IEEE} Conference on Computer Vision and Pattern
  Recognition, {CVPR} 2016, Las Vegas, NV, USA, June 27-30, 2016},  770--778.

\bibitem[\protect\citeauthoryear{Leino and Fredrikson}{2019}]{stolen_memorize}
Leino, K., and Fredrikson, M.
\newblock 2019.
\newblock Stolen memories: Leveraging model memorization for calibrated
  white-box membership inference.
\newblock {\em CoRR} abs/1906.11798.

\bibitem[\protect\citeauthoryear{Li \bgroup et al\mbox.\egroup }{2016a}]{psgld}
Li, C.; Chen, C.; Carlson, D.~E.; and Carin, L.
\newblock 2016a.
\newblock Preconditioned stochastic gradient langevin dynamics for deep neural
  networks.
\newblock In {\em Proceedings of the Thirtieth {AAAI} Conference on Artificial
  Intelligence, February 12-17, 2016, Phoenix, Arizona, {USA.}},  1788--1794.

\bibitem[\protect\citeauthoryear{Li \bgroup et al\mbox.\egroup
  }{2016b}]{sgld_shape_classification}
Li, C.; Stevens, A.; Chen, C.; Pu, Y.; Gan, Z.; and Carin, L.
\newblock 2016b.
\newblock Learning weight uncertainty with stochastic gradient {MCMC} for shape
  classification.
\newblock In {\em 2016 {IEEE} Conference on Computer Vision and Pattern
  Recognition, {CVPR} 2016, Las Vegas, NV, USA, June 27-30, 2016},  5666--5675.

\bibitem[\protect\citeauthoryear{Long \bgroup et al\mbox.\egroup
  }{2018}]{member_gen}
Long, Y.; Bindschaedler, V.; Wang, L.; Bu, D.; Wang, X.; Tang, H.; Gunter,
  C.~A.; and Chen, K.
\newblock 2018.
\newblock Understanding membership inferences on well-generalized learning
  models.
\newblock {\em CoRR} abs/1802.04889.

\bibitem[\protect\citeauthoryear{Mou \bgroup et al\mbox.\egroup
  }{2018}]{colt_sgld_gen}
Mou, W.; Wang, L.; Zhai, X.; and Zheng, K.
\newblock 2018.
\newblock Generalization bounds of {SGLD} for non-convex learning: Two
  theoretical viewpoints.
\newblock In {\em Conference On Learning Theory, {COLT} 2018, Stockholm,
  Sweden, 6-9 July 2018.},  605--638.

\bibitem[\protect\citeauthoryear{Neal and others}{2011}]{langevin_dynamics}
Neal, R.~M., et~al.
\newblock 2011.
\newblock Mcmc using hamiltonian dynamics.
\newblock {\em Handbook of markov chain monte carlo} 2(11):2.

\bibitem[\protect\citeauthoryear{Pensia, Jog, and Loh}{2018}]{ISIT_sgld_gen}
Pensia, A.; Jog, V.; and Loh, P.
\newblock 2018.
\newblock Generalization error bounds for noisy, iterative algorithms.
\newblock In {\em 2018 {IEEE} International Symposium on Information Theory,
  {ISIT} 2018, Vail, CO, USA, June 17-22, 2018},  546--550.

\bibitem[\protect\citeauthoryear{Robbins and
  Monro}{1951}]{stochastic_optimization}
Robbins, H., and Monro, S.
\newblock 1951.
\newblock A stochastic approximation method.
\newblock {\em The annals of mathematical statistics}  400--407.

\bibitem[\protect\citeauthoryear{Sablayrolles \bgroup et al\mbox.\egroup
  }{2019}]{optimal_attack_ICML}
Sablayrolles, A.; Douze, M.; Schmid, C.; Ollivier, Y.; and Jegou, H.
\newblock 2019.
\newblock White-box vs black-box: Bayes optimal strategies for membership
  inference.
\newblock In {\em Proceedings of the 36th International Conference on Machine
  Learning, {ICML} 2019, 9-15 June 2019, Long Beach, California, {USA}},
  5558--5567.

\bibitem[\protect\citeauthoryear{Sato and Nakagawa}{2014}]{sde_view_sgld}
Sato, I., and Nakagawa, H.
\newblock 2014.
\newblock Approximation analysis of stochastic gradient langevin dynamics by
  using fokker-planck equation and ito process.
\newblock In {\em Proceedings of the 31th International Conference on Machine
  Learning, {ICML} 2014, Beijing, China, 21-26 June 2014},  982--990.

\bibitem[\protect\citeauthoryear{Shalev{-}Shwartz \bgroup et al\mbox.\egroup
  }{2010}]{shwartz_stable}
Shalev{-}Shwartz, S.; Shamir, O.; Srebro, N.; and Sridharan, K.
\newblock 2010.
\newblock Learnability, stability and uniform convergence.
\newblock {\em Journal of Machine Learning Research} 11:2635--2670.

\bibitem[\protect\citeauthoryear{Shokri \bgroup et al\mbox.\egroup
  }{2017}]{s_and_p_attack}
Shokri, R.; Stronati, M.; Song, C.; and Shmatikov, V.
\newblock 2017.
\newblock Membership inference attacks against machine learning models.
\newblock In {\em 2017 {IEEE} Symposium on Security and Privacy, {SP} 2017, San
  Jose, CA, USA, May 22-26, 2017},  3--18.

\bibitem[\protect\citeauthoryear{Wang, Fienberg, and
  Smola}{2015}]{privacy_for_free_icml}
Wang, Y.; Fienberg, S.~E.; and Smola, A.~J.
\newblock 2015.
\newblock Privacy for free: Posterior sampling and stochastic gradient monte
  carlo.
\newblock In {\em Proceedings of the 32nd International Conference on Machine
  Learning, {ICML}, Lille, France, 6-11 July 2015},  2493--2502.

\bibitem[\protect\citeauthoryear{Welling and Teh}{2011}]{sgld_original}
Welling, M., and Teh, Y.~W.
\newblock 2011.
\newblock Bayesian learning via stochastic gradient langevin dynamics.
\newblock In {\em Proceedings of the 28th International Conference on Machine
  Learning, {ICML} 2011, Bellevue, USA, June 28 - July 2, 2011},  681--688.

\bibitem[\protect\citeauthoryear{Wu \bgroup et al\mbox.\egroup
  }{2019a}]{wu2019p3sgd}
Wu, B.; Zhao, S.; Sun, G.; Zhang, X.; Su, Z.; Zeng, C.; and Liu, Z.
\newblock 2019a.
\newblock P3sgd: Patient privacy preserving sgd for regularizing deep cnns in
  pathological image classification.
\newblock In {\em Proceedings of the IEEE Conference on Computer Vision and
  Pattern Recognition},  2099--2108.

\bibitem[\protect\citeauthoryear{Wu \bgroup et al\mbox.\egroup
  }{2019b}]{GAN_privacy}
Wu, B.; Zhao, S.; Xu, H.; Chen, C.; Wang, L.; Zhang, X.; Sun, G.; and Zhou, J.
\newblock 2019b.
\newblock Generalization in generative adversarial networks: {A} novel
  perspective from privacy protection.
\newblock {\em CoRR} abs/1908.07882.

\bibitem[\protect\citeauthoryear{Yeom \bgroup et al\mbox.\egroup
  }{2018}]{overfitting_privacy}
Yeom, S.; Giacomelli, I.; Fredrikson, M.; and Jha, S.
\newblock 2018.
\newblock Privacy risk in machine learning: Analyzing the connection to
  overfitting.
\newblock In {\em 31st {IEEE} Computer Security Foundations Symposium, {CSF}
  2018, Oxford, United Kingdom, July 9-12, 2018},  268--282.

\bibitem[\protect\citeauthoryear{Zagoruyko and Komodakis}{2016}]{wider-resnet}
Zagoruyko, S., and Komodakis, N.
\newblock 2016.
\newblock Wide residual networks.
\newblock In {\em Proceedings of the British Machine Vision Conference 2016,
  {BMVC} 2016, York, UK, September 19-22, 2016}.

\end{thebibliography}

\end{document}